\documentclass[conference]{IEEEtran}
\IEEEoverridecommandlockouts

\usepackage{algorithm}
\usepackage{algpseudocode}
\usepackage{cite} 
\usepackage{amsmath}
\usepackage{comment} 
\usepackage[table,xcdraw]{xcolor}
\usepackage[section]{placeins} 
\usepackage[pdftex]{graphicx}

\title{\LARGE \bf
3D Radar and Camera Co-Calibration: A flexible and Accurate Method for Target-based Extrinsic Calibration
}

\author{\IEEEauthorblockN{Lei Cheng, Arindam Sengupta and Siyang Cao}
\IEEEauthorblockA{Department of Electrical and Computer Engineering, University of Arizona, Tucson, AZ, USA\\
Email: leicheng@arizona.edu, sengupta@arizona.edu, caos@arizona.edu}
}

\begin{document}

\maketitle
\thispagestyle{empty}
\pagestyle{empty}

\begin{abstract}
Advances in autonomous driving are inseparable from sensor fusion. Heterogeneous sensors are widely used for sensor fusion due to their complementary properties, with radar and camera being the most equipped sensors. Intrinsic and extrinsic calibration are essential steps in sensor fusion. The extrinsic calibration, independent of the sensor's own parameters, and performed after the sensors are installed, greatly determines the accuracy of sensor fusion. Many target-based methods require cumbersome operating procedures and well-designed experimental conditions, making them extremely challenging. To this end, we propose a flexible, easy-to-reproduce and accurate method for extrinsic calibration of 3D radar and camera. The proposed method does not require a specially designed calibration environment, and instead places a single corner reflector (CR) on the ground to iteratively collect radar and camera data simultaneously using Robot Operating System (ROS), and obtain radar-camera point correspondences based on their timestamps, and then use these point correspondences as input to solve the perspective-n-point (PnP) problem, and finally get the extrinsic calibration matrix. Also, RANSAC is used for robustness and the Levenberg-Marquardt (LM) nonlinear optimization algorithm is used for accuracy. Multiple controlled environment experiments as well as real-world experiments demonstrate the efficiency and accuracy (AED error is 15.31 pixels and Acc up to 89\%) of the proposed method.
\end{abstract}
\vspace{-0.1cm}
\section{INTRODUCTION}
Self-driving cars and mobile robots, which are hailed as new applications that could change the world, are gradually becoming a reality. They use multiple types of sensors for high-quality perception and decision-making, of which camera and radar are the most commonly used.
These sensors inherently have their own defects and merits due to their physical operating principles\cite{liu2021robust,wang2021rethinking}.
Thus a heterogeneous sensor fusion system, which is able to exploit the complementary properties of different sensors, is required to achieve accurate, robust and reliable sensing capabilities. 
Radar and camera extrinsic calibration is crucial for the sensor fusion between these two sensors. It directly associates the radar point clouds with the corresponding visual targets so as to make full use of the complementary data of heterogeneous sensors to truly play the role of sensor fusion.
\begin{figure*} 
	\centering
	\includegraphics[width=0.98\textwidth]{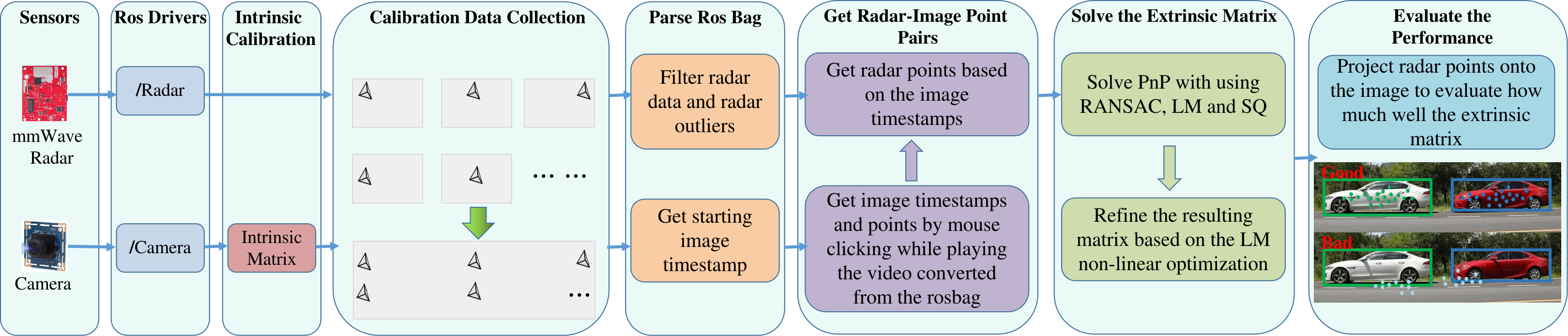}
	\vspace{-0.15cm}
	\caption{The overall pipeline of the proposed radar-camera extrinsic co-calibration method.}
	\label{10}
	\vspace{-0.35cm}
\end{figure*}
Extrinsic calibration estimates the spatial transformation of the sensor coordinates to other sensors or unified reference frames\cite{domhof2019extrinsic}. It requires finding feature correspondences perceived by both sensors. Target-based methods can achieve the higher accuracy than targetless methods with the dedicated calibration targets and deliberate calibration procedures\cite{tsai2021optimising,pervsic2019extrinsic}. High association accuracy is the ultimate goal of calibration, so our research is built on target-based methods.
However, extrinsic calibration involving radar is always a tough task. First, radar detections are noisy and unstable due to multipath reflections, jitters and inconsistency\cite{wise2021continuous}. Second, the calibration process is tedious, time-consuming and cumbersome. 

Recently, there have been numerous approaches proposed for lidar and camera extrinsic calibration, while been few for radar and camera. Simply adapting the methods of those lidars to radars is not reasonable because the two sensors work significantly differently. Many papers on sensor fusion lack procedural descriptions of extrinsic calibration or employ very simple calibration methods\cite{sengupta2022automatic,bai2021robust,fu2020camera}, such as estimating calibration parameter by measuring the installation position and angle between the radar and camera, but the calibration accuracy suffers from these imprecise or even impossible instrumental measurements especially for angle measurements. This is partly because most works primarily focus on sensor fusion algorithms than calibration, and partly because no state-of-the-art and easy-to-use calibration method exists.
Existing calibration methods use various custom calibration targets, such as using multiple corner reflectors and checkerboards, metal poles or balls, which affects the reproducibility of these methods. Some methods involve the use of support stands to adjust the heights of the corner reflectors or the need to move the corner reflector, support stands themselves and the motion of the corner reflector, which can interfere with radar detection, are potential factors that lead to inaccurate calibration\cite{olutomilayo2021extrinsic,pervsic2019extrinsic,domhof2019extrinsic,oh2018comparative,sugimoto2004obstacle,el2015radar}. These methods also require intensive operating procedures.
In addition, deep learning has been highly sought after in radar-camera sensor fusion, which needs accurate extrinsic calibration to generate well-labeled data-sets\cite{zhang2021rvdet,dong2021radar}. Some calibration methods even requires lidar to provide ground truth data, which is cost intensive.
Most importantly, the vast majority of existing calibration methods only consider 2D radar detection by either using a 2D radar board or constraining the radar detection to a 2D plane. Therefore, they are not suitable for the promising 3D radar related extrinsic calibration.

Based on the shortcomings of existing methods and the scarcity of radar-camera extrinsic calibration methods, and inspired by the 3D radar target-less extrinsic calibration presented in \cite{wise2021continuous}, we propose a target-based 3D radar-camera calibration method that is simple to operate, easy to reproduce and accurate.
The method uses only a single CR as the calibration target, which not only avoids radar reflection interference caused by multiple calibration targets or the tripods that support them, but also facilitates experimental reproduction by other researchers with minimal effort. It performs outliers filtering to alleviate radar detection instability, and adopts RANSAC and the LM optimization algorithm to make the calibration robust and accurate. The overall pipeline is shown in Fig. \ref{10}.

The rest of this paper is organized as follows. The problem statement and formulation is described in sections II. The proposed method is elaborated in section III. The experimental setup and results are presented and discussed in section IV followed by the conclusion.
\vspace{0.1cm}
\section{PROBLEM STATEMENT}
\vspace{-0.1cm}
\subsection{Radar-Camera Coordinate System Transformation} 
\begin{figure}[h!]
	\centering
	\includegraphics[width=0.48\textwidth]{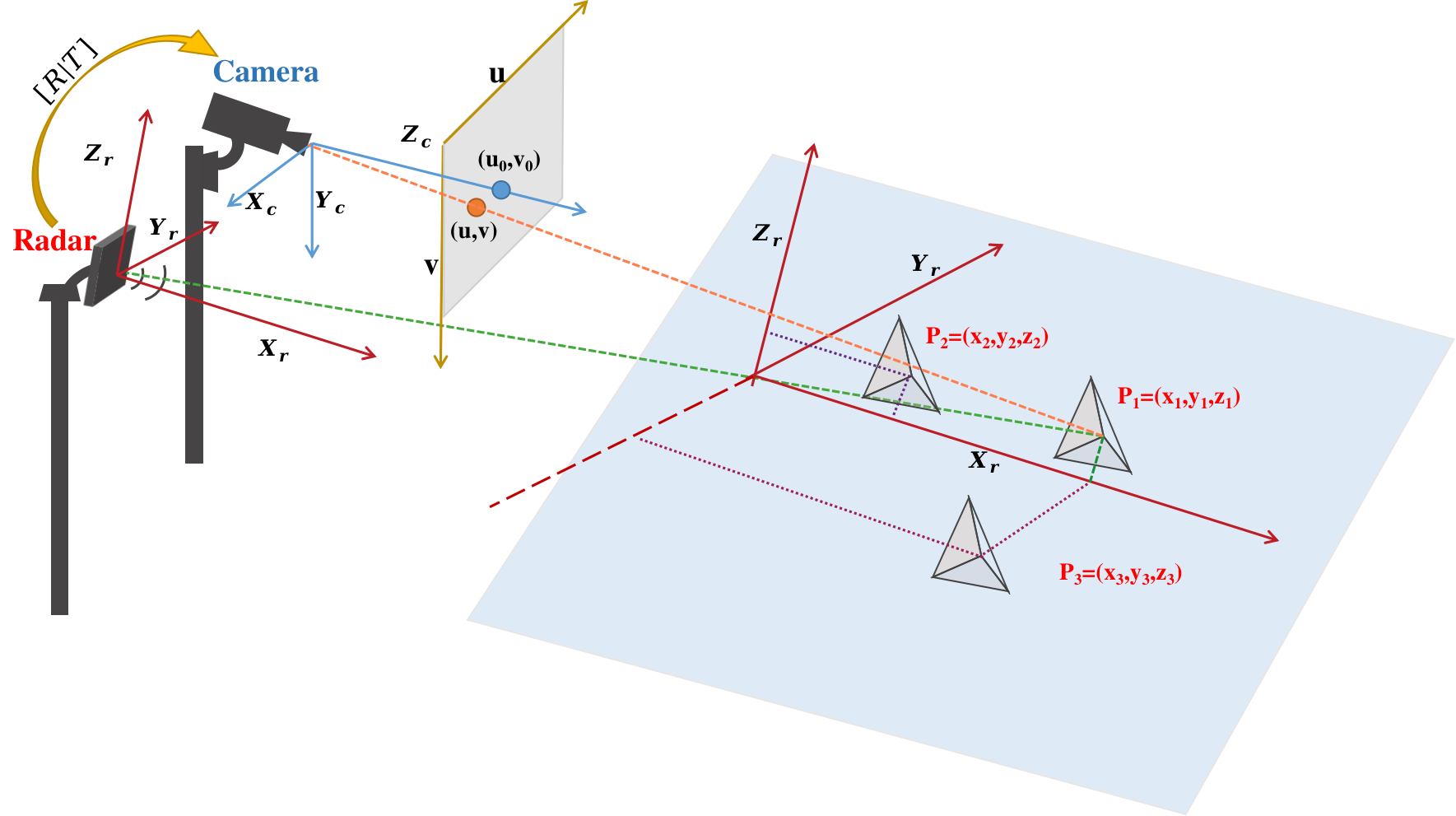}
	\vspace{-0.25cm}
	\caption{Radar-camera coordinate system transformation and data collection demonstrations. $P_1$, $P_2$, and $P_3$ are the three positions where a CR is placed asynchronously three times, respectively.}
	\label{7}
	\vspace{-0.05cm}
\end{figure}

Calibration can be referred to as solving a transformation matrix that can transform a point in the world coordinate system (WCS) to the sensor coordinate system (SCS), as depicted in Fig. \ref{7}. This transformation matrix consists of the intrinsic matrix and the extrinsic matrix. The intrinsic camera calibration is to estimate a matrix (denoted as $K$), which can map a point $P_c=(X_c,Y_c,Z_c)$ in the 3D camera coordinate system (CCS) to its corresponding point $P_p=(u,v)$ in the 2D pixel coordinate system (PCS)\cite{zhang2000flexible}. 
Radar-camera extrinsic calibration can be thought of as determining the rigid body transformation from the camera coordinate system to the radar coordinate system (RCS). Taking the RCS as the WCS, this rigid transformation can be defined as:
\vspace{-0.1cm}
\begin{equation}\label{eq6}
P_c  =  [R|T]\ P_r
\end{equation}
where $P_r=(X_r,Y_r,Z_r)$ is a 3D point in the RCS. Note that points should be in homogeneous coordinates for real calculation. 
Further, the transformation between the RCS and the PCS can be expressed as:
\vspace{-0.1cm}
\begin{equation}\label{eq7}
s\ P_p  = K\ [R|T]\ P_r =K Q P_r
\end{equation}
where $s$ is the scale factor,$R$ is a 3-by-3 rotation matrix, and $T$ is a 3-by-1 translation vector. The extrinsic matrix, denoted as Q, can be represented as $[R|T]$.
In addition, the rotation matrix of the rigid body transformation can be expressed as a sequence of three rotations, one about each principle axis\cite{slabaugh1999computing}. Considering the $xyz$ sequence of rotations, which correspond to the Euler angles, $\psi$, $\theta$, and $\phi$ respectively, the rotation matrix $R$ can be rewritten as:
\vspace{-0.1cm}
\begin{equation}\label{eq8}
R  = R_z(\phi) R_y(\theta) R_x(\psi)
\end{equation}
Therefore, the radar-camera extrinsic calibration is to solve a 6 DoF rigid body transformation which can be represented with the three translation parameters,$[t_x,t_y,t_z]$, and three rotation parameters,$[\psi,\theta,\phi]$.
\subsection{Radar-Camera Extrinsic Calibration Solution}
The objective of the radar-camera extrinsic calibration is to determine the extrinsic matrix based on the known camera intrinsic parameters and a set of N correspondences between 3D points in RCS and their corresponding 2D points in PCS, which is known as the PnP problem.
As we mentioned earlier, the extrinsic matrix has 6 independent parameters, and three points known in both coordinate systems can provide nine constraints which are quite sufficient to solve when the three points are not colinear\cite{horn1987closed}. In reality, the measurements of these two sensors are both not exact and they may not match well, so to achieve high accuracy, more than 3 points are needed.
When more than 3 points are used, the calibration problem is transformed into a task aimed at minimizing the sum of squares of residual errors between the projected radar 2D points and the corresponding 2D ground truth pixel points. And it is intrinsically a nonlinear problem.

The LM algorithm, which combines the advantages of steepest descent and the Gauss-Newton method, is the ``gold-standard" solution to the nonlinear problem\cite{marchand2015pose}. 
However, the algorithm requires a good initial guess to converge to the globally optimal solution\cite{marchand2015pose}. RANSAC, as the robust solution that can account for noisy radar measurements, can serve as the initial guess. Radar measurements are always noisy or perturbed that the radar-camera point correspondences are always inexact, which can even be fatal for high-quality calibration processes that require perfect matches.
RANSAC is an iterative method, which uses the smallest set of possible correspondences to estimate parameters of a mathematical model, and rejects outliers. In turn, it can well identify mismatches and provide a robust solution that can address noise. However, RANSAC always uses a minimal number of correspondences to gather a set of inliers, which leads to its performance suffers from a highly contaminated measurement set. Accordingly, the RANSAC process followed by the LM nonlinear optimization process adopted as our solution.
\section{EXTRINSIC CALIBRATION METHOD}
\subsection{Calibration Target Choice}
The choice of calibration target is undeniably important for target-based calibration methods. For accurate radar-camera calibration, the calibration target should provide unbiased return signals as much as possible to ensure localization accuracy, and on the other hand, should provide as high a reflection intensity as possible to ensure target visibility. Therefore, Innovative Technical Systems' trihedral corner reflector TCR-46, which has a radar cross-section of 1.23 dBsm, was chosen as the calibration target to serve both purposes. 
\vspace{-0.05cm}
\renewcommand{\algorithmiccomment}[1]{\hfill$\triangleright$\textit{\textcolor{blue}{#1}}}
\begin{algorithm}
\small
\caption{Estimating the Radar-Camera Extrinsic Matrix}\label{alg1}
\textbf{Input: }Radar rosbag, Camera rosbag, Intrinsic Matrix($K$), Distortion coefficents($D$)\\
\textbf{Output: }Extrinsic Matrix($Q$)
\begin{algorithmic}[1]
\State Convert camera rosbag to video file
\State Play the video and mouse click on the center of the CR in the image to get its pixel coordinates($img\_2d$) and timestamps($local\_ts$).
\State Get starting timestamp $start\_ts$ from camera rosbag and radar timestamps $rad\_ts$ from radar rosbag
\State $img\_ts \gets local\_ts+start\_ts$  \Comment{Get image timestamps}
\State $Rad \gets Rad[(velocity = 0) \& (range<20)]$  \Comment{Filter radar data}
\State $rad2img\_ts \gets find\_closest\_ts(rad\_ts, img\_ts)$ \Comment{Find the closest radar timestamps to the image timestamps}

\For{$ts \in rad2img\_ts$}
\State $[x,y,z] \gets Rad[rad\_ts \ge ts-1) \& (rad\_ts \le ts+1)]$

                     \Comment{Get radar data within 3 secs with ts as an integer}
\State $cond \gets {(|Zscore(x)| < thr\_x)}\& (|Zscore(y)| < thr\_y)\& (|Zscore(z)| < thr\_z)$ \Comment{Radar outliers condition with using Z-score}
\If{$cond = 1$}
    \State $rad\_3d.append([x.mean,y.mean,z.mean])$
\EndIf
\EndFor
\State $err\_it,Q\_it \gets solvePnPR\_it(rad\_3d,img\_2d,K,D)$

                     \Comment{Solve PnP with RANSAC based on the iterative optimization}
\State $err\_sq,Q\_sq \gets solvePnPR\_sq(rad\_3d,img\_2d,K,D)$ 

                     \Comment{Solve PnP with RANSAC based on the quadratically constrained quadratic program} 
\If{$err\_it < err\_sq$}
    \State $Q \gets Q\_it$
\Else
    \State $Q \gets Q\_sq$
\EndIf
\State $Q \gets ExtrinsicsRefineLM(Q,rad\_3d,img\_2d,K,D)$ 

                     \Comment{Refine the extrinsic matrix based on the non-linear LM minimization}
\State Reprojection with $Q$ for performance evaluation
\end{algorithmic}
\vspace{-0.1cm}
\end{algorithm}
\vspace{-0.25cm}
\subsection{Data Collection}
\vspace{-0.1cm}
The accuracy of the calibration is highly dependent on the quality of the data used for the calibration. Therefore, data collection is an important process that cannot be overemphasized.
We use a 9 x 6 checkerboard pattern with a 22.14 mm square edge size as our calibration board to collect data for camera intrinsic calibration by following the Zhang's method\cite{zhang2000flexible}. The ROS camera calibrator were used to estimate the intrinsic parameters.
Radar intrinsic calibration is usually done at the factory using a known point-target in the anechoic chamber, and its intrinsic parameters can be considered a priori. 

For radar-camera extrinsic calibration data collection, first, we fix a radar and a camera on a tripod and tilt the radar and camera slightly so that they look down at the ground. Then, a CR is placed on the ground within the common field of view(FoV) of the radar and camera. Because the experimental field is open and there is not much clutter from other objects, the reflected signal of the CR is conspicuous. Next, we change the placement location of the CR each time and repeat to get enough data. Using one CR multiple times ($N$ times) can simulate placing any number ($N$ corner reflectors) of corner reflectors at the same time. Finally, with the ROS, the sensory data of radar and camera is time stamped and recorded in ROS bags in real time.

In this study, we aim to perform full 6 DoF 3D radar-camera calibration. A TI AWR1843 3D radar unit with available elevation channels is adopted, and its detections are converted from polar representation to 3D Cartesian coordinate representation. The data collection procedures we mentioned above allow us to collect data with different elevations, as illustrated in Fig. \ref{7}.
In the RCS, $z$ is vertically upward along the radar board, $x$ and $y$ are in a plane perpendicular to the radar board, $x$ is normal to the radar board and points forward, and $y$ is to the right of $x$ and orthogonal to $x$. When the radar board is tilted, if the CR moves back and forth towards the radar, it will cause the readings on the $x$ and $z$ axes to change simultaneously, and if the CR moves left and right parallel to the radar, it will cause readings on the $x$ and $y$ axes to change. Valid 3D radar data will be collected by placing a CR on the ground toward and parallel to the radar multiple times.
Instead of using a height-adjustable tripod to place the CR, we can collect 3D data within a limited range by simply moving the CR freely on the ground, and without having to adjust the height of the tripod from time to time.
\subsection{Data Processing}
The collected data of radar and camera are stored in two separate ROS bags, which can be parsed by their corresponding ROS topics. First, we developed a program that converts the camera's ROS bag into a video, and then while playing that video, each time the mouse clicks on the center of the CR, the program automatically generates the pixel position of the corresponding CR and the corresponding timestamp. Then, we use these timestamps (stored in numpy array) to find the corresponding radar points in the radar's ROS bag that are closest to the timestamps.
However, given the volatility of radar points, the radar point that best corresponds to the image timestamp is still not reliable. We aggregate all radar points within three seconds (considering the frame rate of $K$ Hz, the total is $3K$ frames of points) centered on the image timestamp, and filter the $x$, $y$, and $z$ axis data separately to get points that can be used for calibration (i.e., inliers), and then take the average of these inliers as the final radar points. The Z-Score, is defined as a measure of the divergence of a dataset from the mean, is being used in the filtering process to prune out outliers. We combine these obtained image pixel points (stored as numpy arrays) and radar points (stored as numpy arrays) one by one to generate sets of radar-camera point pairs. Besides, radar points can be filtered based on whether the calibration target is static and its placement range.

With the pre-obtained intrinsic matrix, we can perform joint calibration based on the radar-camera point pairs. The RANSAC will be used for ruling out the unsuitable radar-camera point pairs which may due to a certain time CR's placement. These RANSAC-passed point pairs are used as input to solve the PnP problem using an iterative method based on an LM optimization, followed by a globally optimal method SQPnP. SQPnP casts the PnP problem as a quadratically constrained quadratic program and solves it by conducting local searches in the vicinity of special feasible points from which the global minima are always determined\cite{terzakis2020consistently}. The one with the smaller reprojection error of the two calibration matrices obtained by the two methods, will then be fed into the refinement process. The refinement step still uses the LM minimization scheme, and all radar-camera point pairs including outliers will be used in this step to account for the shortcoming of RANSAC as pointed out in Section II and thus further improve the calibration accuracy. All the aforementioned steps are presented in Algorithm \ref{alg1}. 
\section{EXPERIMENTAL RESULTS AND DISCUSSION}
\subsection{Experimental Setup}
\begin{figure}[h!]
	\centering
	\includegraphics[width=0.4\textwidth]{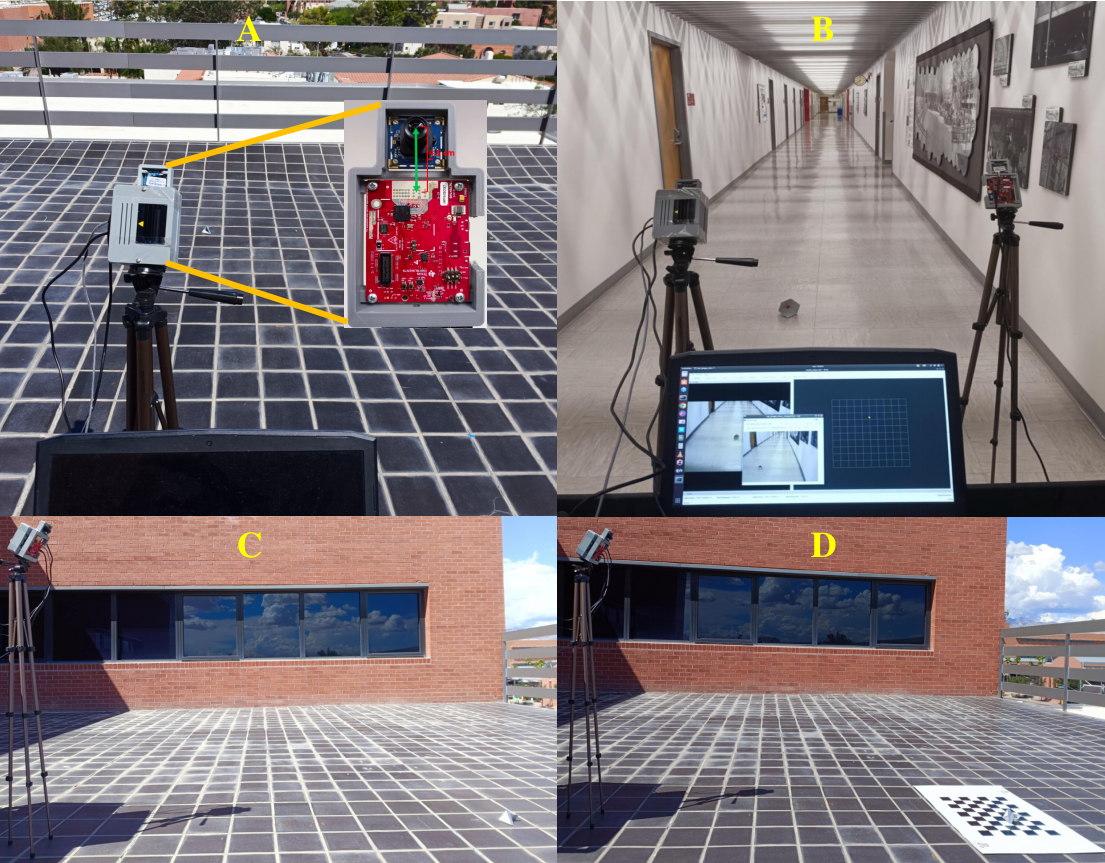}
	\vspace{-0.25cm}
	\caption{Experimental setup and scenarios: (A) The radar-camera data collection setup with both sensors are mounted on the same rig and the vertical distance between them measures 4.5cm; (B) The data collection in the aisle with two sensors are mounted on different rigs; (C) The data collection scenario for PPM; (D) The data collection scenario for CCM.}
	\label{1}
	\vspace{-0.25cm}
\end{figure}

We use a TI AWR1843 3D mmWave radar and a USB8MP02G monocular camera as our two sensing modalities.  They are mounted on a 3D printed rig and are connected with the laptop running Ubuntu 18 and ROS Melodic, with the sensor system synchronized using a common ROS time server. The experimental setup is shown in Fig. \ref{1}.
\subsection{Evaluation Metrics}
In order to evaluate the calibration accuracy, we provide three evaluation metrics: the average euclidean distance(AED), the corrected distance standard deviation(CDSD) and the radar-camera association accuracy (Acc). They are defined as below:
\vspace{-0.1cm}
{\small
\begin{align}
AED &= \frac{\sum_{i=1}^{N}ED_{i}}{N} \label{a}\\
CDSD &= \sqrt{\frac{\sum_{i=1}^{N}(ED_{i}-AED)^{2}}{N-1}} \label{b}\\
Acc(\%) &= \frac{\textit{No. of radar points inside Image BBox}}{\textit{Total no. of radar detections}} \label{c}
\end{align}
\vspace{-0.1cm}
}
where $N$ is the number of point pairs, $ED$ is the euclidean distance for one pair of points.
AED evaluates the distances between the original image points and the corresponding radar-projected points. CDSD evaluates the deviation of these distances. Acc is used to determine the fraction of projected radar points for a given target that accurately falls inside its image bounding box(BBox).
In addition to the above objective evaluation metrics, for simplicity and intuitiveness, we also choose to use the subjective comparison method, that is, to project radar points into camera images and check how well the projected points match the positions of the targets by human eyes.
\begin{table}[h!]
\centering
\vspace{-0.25cm}
\caption{The performance of the co-calibration methods.}
\label{tab1}
\resizebox{0.3\textwidth}{!}{%
\begin{tabular}{|l||l|l|}
\hline
\rowcolor[HTML]{73A9AD} 
\multicolumn{1}{|c||}{\cellcolor[HTML]{ACE1AF}\textbf{Methods}} & \textbf{AED} & \textbf{CDSD} \\ \hline
\rowcolor[HTML]{F5F0BB} 
\cellcolor[HTML]{C4DFAA}\textbf{PPM+D1}      & 15.31   & 9.40     \\ \hline
\rowcolor[HTML]{F5F0BB} 
\cellcolor[HTML]{C4DFAA}\textbf{PMM+D1}      & 79.38   & 19.08    \\ \hline
\rowcolor[HTML]{F5F0BB} 
\cellcolor[HTML]{C4DFAA}\textbf{CCM+D2}      & 7.38    & 4.39     \\ \hline
\rowcolor[HTML]{F5F0BB} 
\cellcolor[HTML]{C4DFAA}\textbf{PPM+D2}      & 16.87   & 6.37     \\ \hline
\rowcolor[HTML]{F5F0BB} 
\cellcolor[HTML]{C4DFAA}\textbf{CCM+D1}      & 8935.97 & 14403.44 \\ \hline
\rowcolor[HTML]{F5F0BB} 
\cellcolor[HTML]{C4DFAA}\textbf{PPM+D1+T\_m} & 15.86   & 9.07     \\ \hline
\rowcolor[HTML]{F5F0BB} 
\cellcolor[HTML]{C4DFAA}\textbf{PMM+D1+T\_c} & 87.07   & 18.62    \\ \hline
\rowcolor[HTML]{F5F0BB} 
\cellcolor[HTML]{C4DFAA}\textbf{PPM+D3}      & 11.03   & 8.18     \\ \hline
\rowcolor[HTML]{F5F0BB} 
\cellcolor[HTML]{C4DFAA}\textbf{PPM+DR+D3}  & 53.90   & 70.13    \\ \hline
\end{tabular}%
}
\vspace{-0.35cm}
\end{table}
\subsection{Results and Discussion}
Multiple controlled environment experiments as well as real-world experiments are performed to demonstrate the efficiency and accuracy of the proposed method.
We choose the physical measurement method(denoted as PMM)\cite{sengupta2022automatic} and the checkerboard combined with the CR method (denoted as CCM) for experimental comparison with the proposed method (denoted as PPM). For the PMM, since the radar and camera are mounted on the same rig in the same plane and they are vertically collinear (Fig. \ref{1}(A)), it can be considered that there is neither a rotation (resulting in an identity rotation matrix) nor a depth and lateral offset between the sensors. The extrinsic matrix can be completely determined just by measuring the vertical offset to get the translation vector. For the CCM, it is just to put the CR on the intersections of the checkerboard (Fig. \ref{1}(D)) to collect both radar and camera data.
\subsubsection{Controlled Environment Experiments}
\begin{figure}[h!]
	\centering
	\vspace{-0.1cm}
	\includegraphics[width=0.44\textwidth]{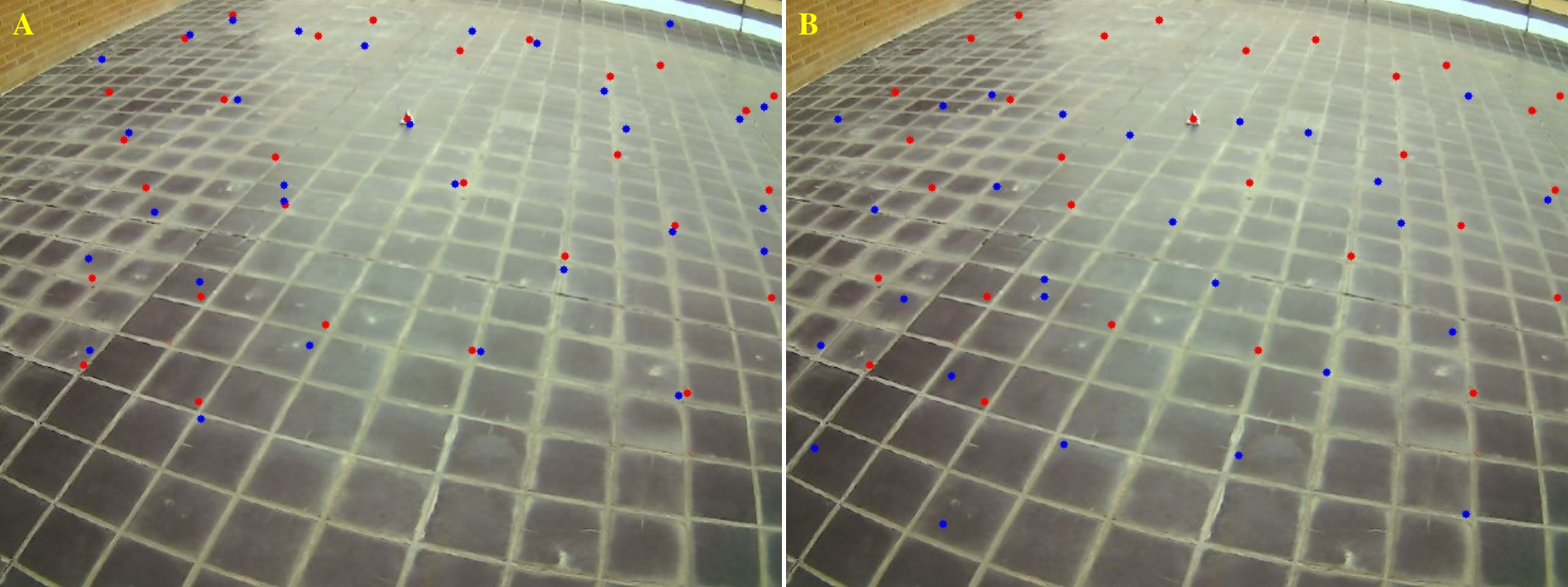}
	\vspace{-0.25cm}
	\caption{The projected radar points and corresponding image points on the ground. The red dots show all CR placement positions while the blue dots show the positions where the radar points are projected onto. (A) PPM+D1; (B) PMM+D1.}
	\label{2}
	\vspace{-0.25cm}
\end{figure}

\begin{figure}[h!]
	\centering
	\vspace{-0.1cm}
	\includegraphics[width=0.45\textwidth]{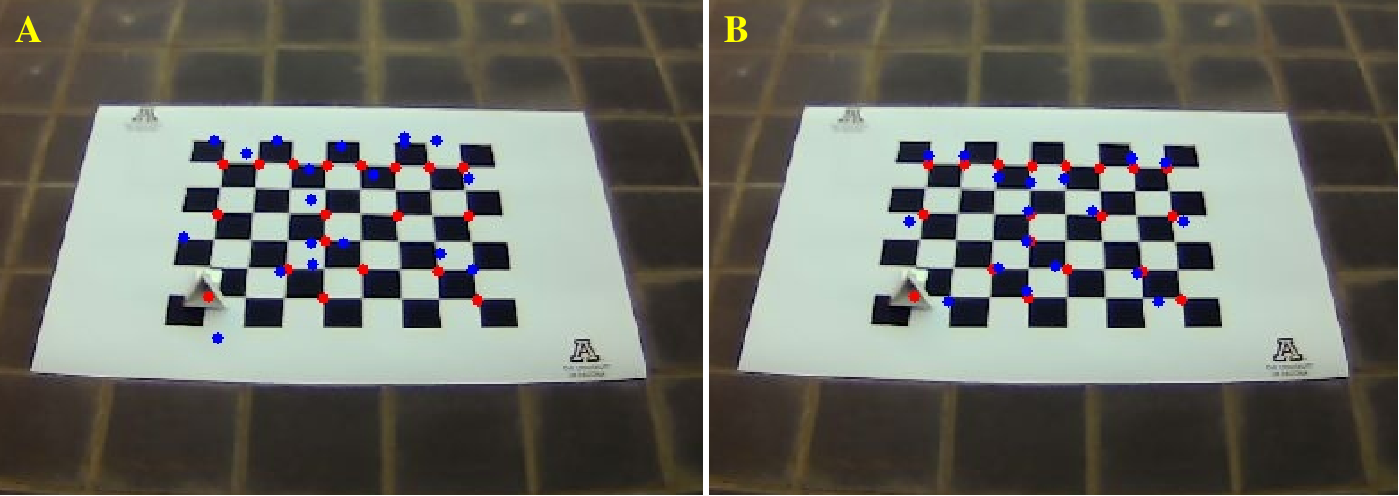}
	\vspace{-0.25cm}
	\caption{The projected radar points and corresponding image points on the checkerboard. (A) PPM+D2; (B) CCM+D2.}
	\label{3}
	\vspace{-0.05cm}
\end{figure}

We collected 2 calibration datasets, D1 and D2, on the rooftop using the PPM method and the CCM method, respectively.
The Fig. \ref{2} illustrates that the image points and projected radar points of the PMM are very well matched compared to the PMM. 
The performance of these methods are presented in Table \ref{tab1}. For the PPM+D1, its AED is 15.31 (which means that the projected point is about 15 pixels away from the original point) and CDSD is 9.40 (which means that the distance-difference distribution is less uneven), both significantly better than the PMM+D1. In order to verify that the poor performance of the PMM is not caused by the inaccuracy of the translation parameter measurement, we use the measured translation vector combined with the rotation matrix obtained by PPM for reprojection, and the results (PPM+D1+T\_m) are almost the same as PPM+D1, and the reprojection performance (PMM+D1+T\_c) using the measured rotation matrix combined with the translation vector obtained by PPM is also almost the same as PMM+D1. This implies that what really plays a huge role is the rotation matrix, which is what PPM aims to estimate. Since the precise physical measurement of the angle of rotation is a big challenge, the PMM is difficult to ensure ideal results. Fig. \ref{3} and Table \ref{tab1} demonstrate that PPM+D2 achieves comparable accuracy to CCM+D2. However, the matrix obtained by the CCM is ill-conditioned. When it is used to reproject D1 data (CCM+D1), the projected points are all outside the image, and its AED and CDSD are abnormally large. The reason may be that the size of the checkerboard is too small, resulting in over-dense data and overfitting.
\par
\begin{figure}[h!]
	\centering
	\vspace{-0.1cm}
	\includegraphics[width=0.42\textwidth]{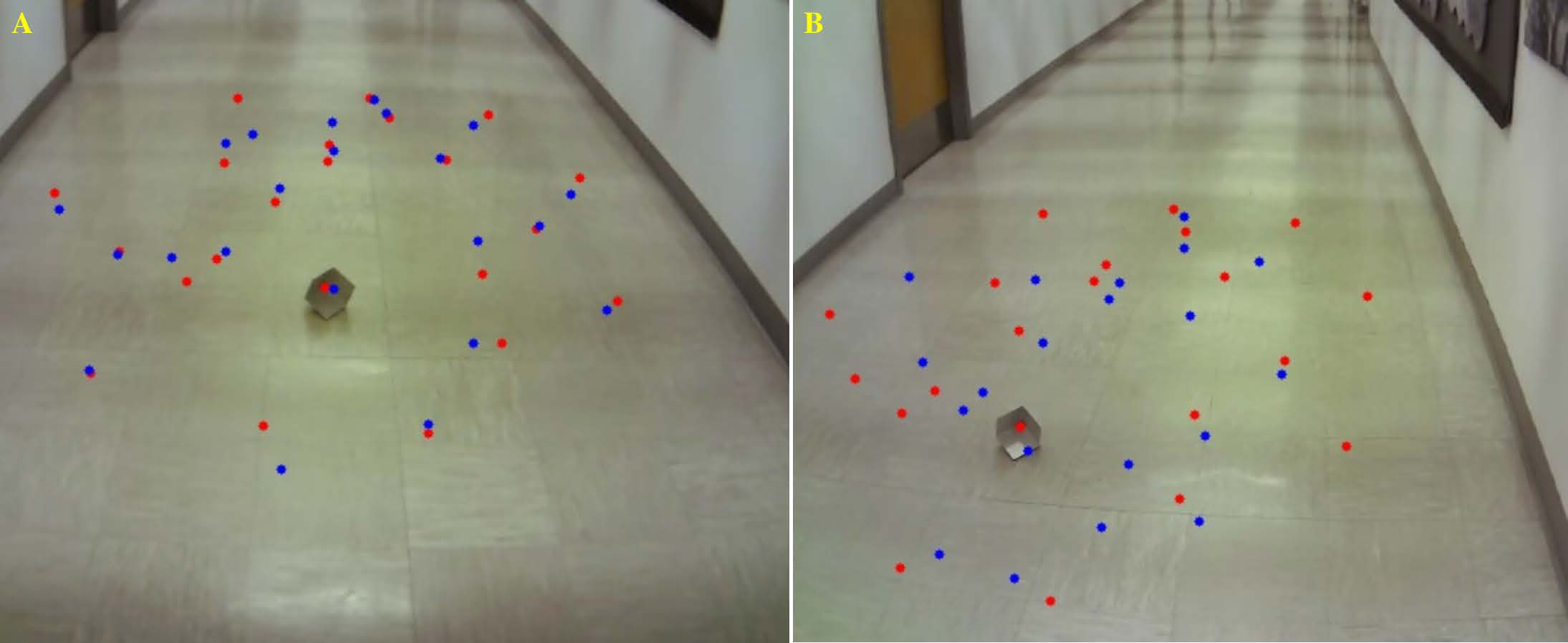}
	\vspace{-0.25cm}
	\caption{The projected radar points and corresponding image points in the aisle. (A) Radar and camera are on the same rig; (B) Radar and camera are on the different rigs.}
	\label{4}
	\vspace{-0.05cm}
\end{figure}

 In addition, we use one radar and two cameras (one on the same rig as the radar and one on a different rig) to collect data (D3) at the same time in the aisle (Fig. \ref{1}(B)) based on the PPM with a TCR-80 CR (larger than the TCR-46) used as the calibration target. The PPM+D3 in Table \ref{tab1} is for camera and radar on the same rig, and its performance data and subjective results (Fig. \ref{4}(A)) show that PPM is still effective even when a larger CR is used and the data collection scene changes. The performance data of PPM+DR+D3 for camera and radar in different rigs is not ideal. However, its subjective results are acceptable. A closer look at Fig. \ref{4}(B) reveals that the distances between several projected points and the original points are very large resulting in a drop in overall performance, which is consistent with its large CDSD.
\subsubsection{Real-World Experiments}
We perform three real-world experiments: people walking and cars moving on the road for the radar and camera on the same rig, and a robot car moving for the radar and camera on the different rigs.
\begin{figure}[h!]
	\centering
	\vspace{-0.1cm}
	\includegraphics[width=0.48\textwidth]{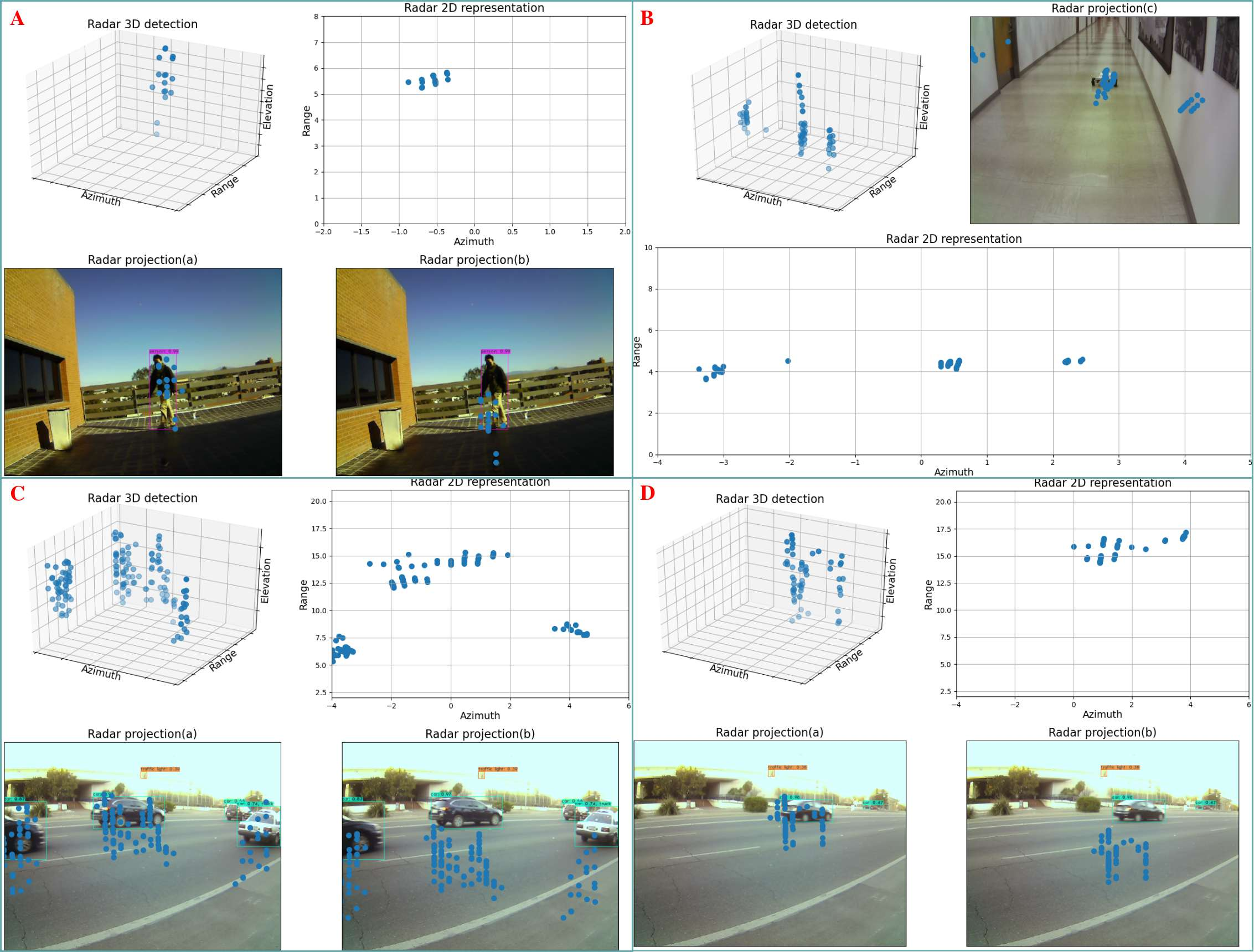}
	\vspace{-0.25cm}
	\caption{Radar projection to the image for real-world experiments: (A) people walking on rooftop; (B) a robot car moving in the aisle; (C) cars moving on the road; (D) a long distance car moving on the road. The Radar projection(a) for the PPM and the Radar projection(b) for the PMM.}
	\label{5}
	\vspace{-0.05cm}
\end{figure}
\begin{figure}[h!]
	\centering
	\vspace{-0.1cm}
	\includegraphics[width=0.43\textwidth]{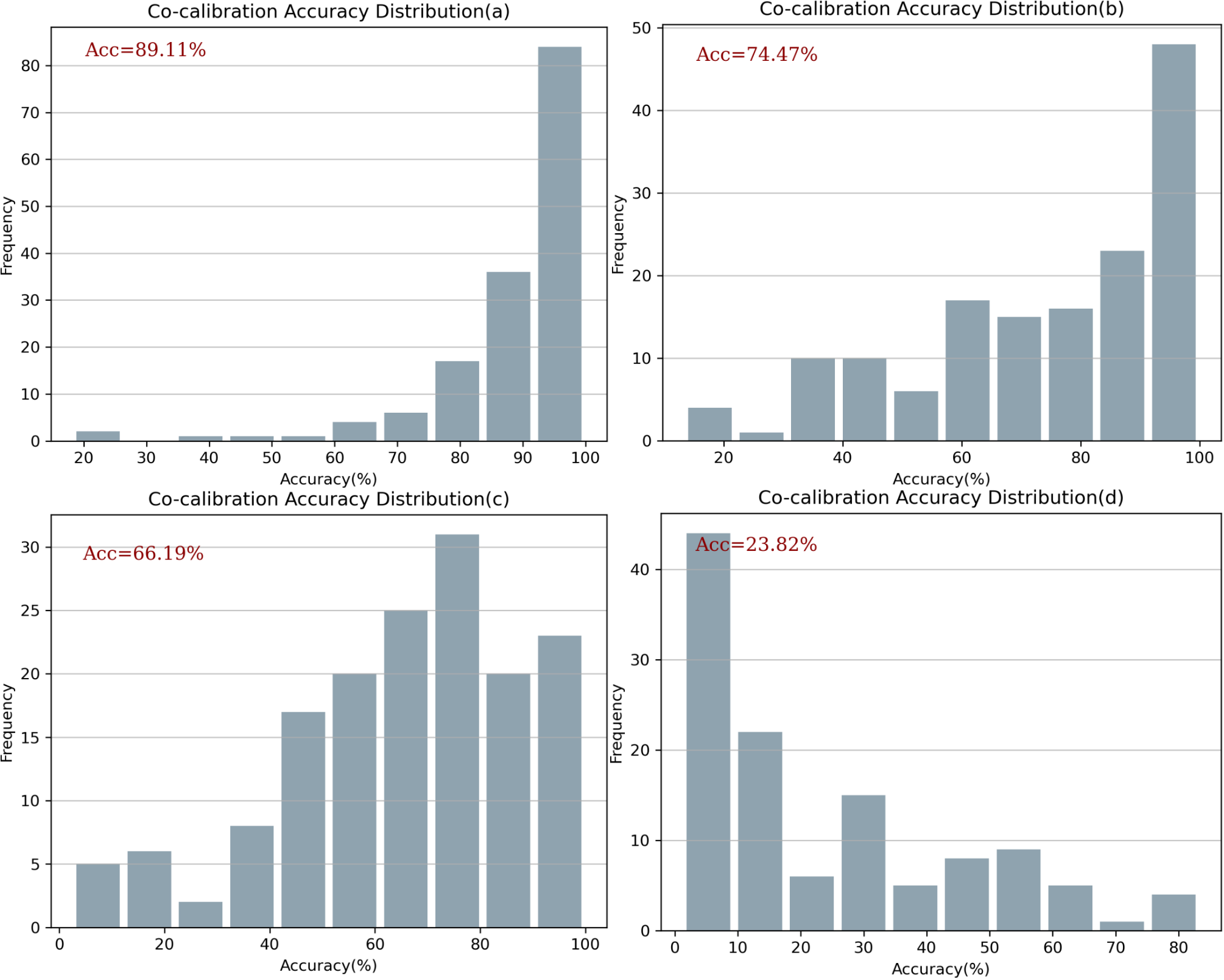}
	\vspace{-0.25cm}
	\caption{Accuracy distribution histograms. (a) The PPM for people walking; (b) The PMM for people walking; (c) The PPM for cars moving; (d) The PMM for cars moving.}
	\label{6}
	\vspace{-0.25cm}
\end{figure}
From Fig. \ref{5}(A, C, D), we can find that most of the projected radar points obtained by PPM fall within the bounding box, which is significantly better than PMM. 
The Fig. \ref{6} provides the histograms of the accuracy distribution of those two methods, from which it still can be found that the PPM is much more accurate than the PMM, especially for moving cars. This may be due to the fact that the car is farther away and traveling at a faster speed, and PMM can't better handle with this situation, as shown in Fig. \ref{5}(D).

At present, most radar-camera co-calibration methods fix the two sensors on the same rig, we perform an experiment of moving a robot car to initially evaluate the proposed method's accuracy with the two sensors on different rigs. As shown in Fig. \ref{5}(B), basically most of the projected radar points fall on the car, although the projection accuracy of the control environment given by PPM+DR+D3 in Table \ref{tab1} is not that ideal. Therefore, the proposed method is suitable for the case where the sensors are not in the same rig.
\vspace{-0.15cm}
\section{CONCLUSIONS}
In this paper, we propose a method for extrinsic calibration of 3D radar and camera. This method uses only a single CR as the calibration target, and there are no strict requirements or regulations for the operation. All operations are to place the CR freely in any open space and repeat it many times to obtain enough data for calibration. This makes the method flexible and easy to reproduce. A series of controlled environmental and real-world experiments, including different experimental scenarios and experimental objects, demonstrate the accuracy and robustness of the proposed method. Specifically, its AED is 15.31 and CDSD is 9.40, while Acc is 89\% for human targets and 66\% for vehicle targets. In the future, we will explore the online calibration method based on the proposed method, and study the method of accurately calibrating the sensors not in the same rig.
\vspace{-0.15cm}
\section*{ACKNOWLEDGMENT}
This work was supported by the Sony Research Award Program. The authors would
like to thank Shuting Hu and Qi Wen for their invaluable help with the experiment.
\addtolength{\textheight}{0cm}   





\vspace{-0.1cm}
\bibliographystyle{ieeetr}
%
{\small
\bibliography{references.bib}}

\begin{thebibliography}{10}

\bibitem{liu2021robust}
Z.~Liu, Y.~Cai, H.~Wang, L.~Chen, H.~Gao, Y.~Jia, and Y.~Li, ``Robust target
  recognition and tracking of self-driving cars with radar and camera
  information fusion under severe weather conditions,'' {\em IEEE Transactions
  on Intelligent Transportation Systems}, 2021.

\bibitem{wang2021rethinking}
Y.~Wang, G.~Wang, H.-M. Hsu, H.~Liu, and J.-N. Hwang, ``Rethinking of radar's
  role: A camera-radar dataset and systematic annotator via coordinate
  alignment,'' in {\em Proceedings of the IEEE/CVF Conference on Computer
  Vision and Pattern Recognition}, pp.~2815--2824, 2021.

\bibitem{domhof2019extrinsic}
J.~Domhof, J.~F. Kooij, and D.~M. Gavrila, ``An extrinsic calibration tool for
  radar, camera and lidar,'' in {\em 2019 International Conference on Robotics
  and Automation (ICRA)}, pp.~8107--8113, IEEE, 2019.

\bibitem{tsai2021optimising}
D.~Tsai, S.~Worrall, M.~Shan, A.~Lohr, and E.~Nebot, ``Optimising the selection
  of samples for robust lidar camera calibration,'' in {\em 2021 IEEE
  International Intelligent Transportation Systems Conference (ITSC)},
  pp.~2631--2638, IEEE, 2021.

\bibitem{pervsic2019extrinsic}
J.~Per{\v{s}}i{\'c}, I.~Markovi{\'c}, and I.~Petrovi{\'c}, ``Extrinsic 6dof
  calibration of a radar--lidar--camera system enhanced by radar cross section
  estimates evaluation,'' {\em Robotics and Autonomous Systems}, vol.~114,
  pp.~217--230, 2019.

\bibitem{wise2021continuous}
E.~Wise, J.~Per{\v{s}}i{\'c}, C.~Grebe, I.~Petrovi{\'c}, and J.~Kelly, ``A
  continuous-time approach for 3d radar-to-camera extrinsic calibration,'' in
  {\em 2021 IEEE International Conference on Robotics and Automation (ICRA)},
  pp.~13164--13170, IEEE, 2021.

\bibitem{sengupta2022automatic}
A.~Sengupta, A.~Yoshizawa, and S.~Cao, ``Automatic radar-camera dataset
  generation for sensor-fusion applications,'' {\em IEEE Robotics and
  Automation Letters}, vol.~7, no.~2, pp.~2875--2882, 2022.

\bibitem{bai2021robust}
J.~Bai, S.~Li, L.~Huang, and H.~Chen, ``Robust detection and tracking method
  for moving object based on radar and camera data fusion,'' {\em IEEE Sensors
  Journal}, vol.~21, no.~9, pp.~10761--10774, 2021.

\bibitem{fu2020camera}
Y.~Fu, D.~Tian, X.~Duan, J.~Zhou, P.~Lang, C.~Lin, and X.~You, ``A
  camera--radar fusion method based on edge computing,'' in {\em 2020 IEEE
  International Conference on Edge Computing (EDGE)}, pp.~9--14, IEEE, 2020.

\bibitem{olutomilayo2021extrinsic}
K.~T. Olutomilayo, M.~Bahramgiri, S.~Nooshabadi, and D.~R. Fuhrmann,
  ``Extrinsic calibration of radar mount position and orientation with multiple
  target configurations,'' {\em IEEE Transactions on Instrumentation and
  Measurement}, vol.~70, pp.~1--13, 2021.

\bibitem{oh2018comparative}
J.~Oh, K.-S. Kim, M.~Park, and S.~Kim, ``A comparative study on camera-radar
  calibration methods,'' in {\em 2018 15th International Conference on Control,
  Automation, Robotics and Vision (ICARCV)}, pp.~1057--1062, IEEE, 2018.

\bibitem{sugimoto2004obstacle}
S.~Sugimoto, H.~Tateda, H.~Takahashi, and M.~Okutomi, ``Obstacle detection
  using millimeter-wave radar and its visualization on image sequence,'' in
  {\em Proceedings of the 17th International Conference on Pattern Recognition,
  2004. ICPR 2004.}, vol.~3, pp.~342--345, IEEE, 2004.

\bibitem{el2015radar}
G.~El~Natour, O.~A. Aider, R.~Rouveure, F.~Berry, and P.~Faure, ``Radar and
  vision sensors calibration for outdoor 3d reconstruction,'' in {\em 2015 IEEE
  International Conference on Robotics and Automation (ICRA)}, pp.~2084--2089,
  IEEE, 2015.

\bibitem{zhang2021rvdet}
J.~Zhang, M.~Zhang, Z.~Fang, Y.~Wang, X.~Zhao, and S.~Pu, ``Rvdet:
  Feature-level fusion of radar and camera for object detection,'' in {\em 2021
  IEEE International Intelligent Transportation Systems Conference (ITSC)},
  pp.~2822--2828, IEEE, 2021.

\bibitem{dong2021radar}
X.~Dong, B.~Zhuang, Y.~Mao, and L.~Liu, ``Radar camera fusion via
  representation learning in autonomous driving,'' in {\em Proceedings of the
  IEEE/CVF Conference on Computer Vision and Pattern Recognition},
  pp.~1672--1681, 2021.

\bibitem{zhang2000flexible}
Z.~Zhang, ``A flexible new technique for camera calibration,'' {\em IEEE
  Transactions on pattern analysis and machine intelligence}, vol.~22, no.~11,
  pp.~1330--1334, 2000.

\bibitem{slabaugh1999computing}
G.~G. Slabaugh, ``Computing euler angles from a rotation matrix,'' {\em
  Retrieved on August}, vol.~6, no.~2000, pp.~39--63, 1999.

\bibitem{horn1987closed}
B.~K. Horn, ``Closed-form solution of absolute orientation using unit
  quaternions,'' {\em Josa a}, vol.~4, no.~4, pp.~629--642, 1987.

\bibitem{marchand2015pose}
E.~Marchand, H.~Uchiyama, and F.~Spindler, ``Pose estimation for augmented
  reality: a hands-on survey,'' {\em IEEE transactions on visualization and
  computer graphics}, vol.~22, no.~12, pp.~2633--2651, 2015.

\bibitem{terzakis2020consistently}
G.~Terzakis and M.~Lourakis, ``A consistently fast and globally optimal
  solution to the perspective-n-point problem,'' in {\em European Conference on
  Computer Vision}, pp.~478--494, Springer, 2020.

\end{thebibliography}

\end{document}